\documentclass[10pt,letterpaper]{article}

\usepackage{graphicx}
\graphicspath{ {./images/} }
\usepackage{multirow}
\usepackage[table,xcdraw]{xcolor}

\renewcommand{\baselinestretch}{1} 

\usepackage{hyperref}
\hypersetup{
    unicode=false,          
    pdftoolbar=true,        
    pdfmenubar=true,        
    pdffitwindow=false,     
    pdfstartview={FitH},    
    pdftitle={GPDer},         
    pdfauthor={Emmanuel Johnson}, 
    pdfsubject={GPDer} 
    pdfcreator={}           
    pdfproducer={},         
    pdfkeywords={},         
    pdfnewwindow=true,      
    colorlinks=true,        
    linkcolor={blue},       
    citecolor=blue,         
    filecolor=blue,         
    urlcolor=blue           
}

\usepackage[colorinlistoftodos,textsize=scriptsize]{todonotes}
\usepackage{flushend}

\usepackage[nocompress]{cite}
\usepackage{tikz}
\usetikzlibrary{arrows,automata,positioning}
\usetikzlibrary{mindmap,trees}

\usepackage{amsmath,amssymb}
\usepackage{amsfonts}
\usepackage{amsbsy} 
\usepackage{bbm} 

\def\x{\ensuremath{\mathbf{x}}}
\def\y{\ensuremath{\mathbf{y}}}

\newcommand{\remove}[1]{}

\usepackage{sections/arxiv}

\usepackage[utf8]{inputenc} 
\usepackage[T1]{fontenc}    
\usepackage{hyperref}       
\usepackage{url}            
\usepackage{booktabs}       
\usepackage{amsfonts}       
\usepackage{nicefrac}       
\usepackage{microtype}      
\usepackage{lipsum}
\usepackage{helvet}
\usepackage{amsmath}

\title{
Accounting for Input Noise in Gaussian Process Parameter Retrieval
}

\author{
  J.~Emmanuel~Johnson\thanks{\url{https://jejjohnson.netlify.app}} \\
  Image Processing Laboratory \\
  Universitat de Val{\`e}ncia\\
  Val{\`e}ncia, Spain\\
  \texttt{juan.johnson@uv.es} \\
  \And
  Valero Laparra\thanks{\url{https://www.uv.es/lapeva/}} \\
  Image Processing Laboratory \\
  Universitat de Val{\`e}ncia\\
  Val{\`e}ncia, Spain\\
  \texttt{valero.laparra@uv.es} \\
  \And
  Gustau Camps-Valls\thanks{\url{https://www.uv.es/gcamps/}} \\
  Image Processing Laboratory \\
  Universitat de Val{\`e}ncia\\
  Val{\`e}ncia, Spain\\
  \texttt{gcamps@uv.es} \\
}

\begin{document}

\begin{center}
    ©IEEE. ACCEPTED FOR PUBLICATION IN IEEE LGRS. DOI 10.1109/LGRS.2019.2921476\footnote{
©IEEE. Personal use of this material is permitted.  Permission from IEEE must be obtained for all other users,including reprinting/republishing this material for advertising or promotional purposes, creating new collectiveworks for resale or redistribution to servers or lists, or reuse of any copyrighted components of this work in otherworks.  DOI: 10.1109/LGRS.2019.2921476.
}
\end{center}

\maketitle



\begin{abstract}
Gaussian Processes (GPs) are a class of kernel methods that have shown to be very useful in geoscience and remote sensing applications for parameter retrieval, model inversion and emulation. They are widely used because they are simple, flexible, and provide accurate estimates. GPs are based on a Bayesian statistical framework which provides a posterior probability function for each estimation. Therefore, besides the usual prediction (given in this case by the mean function), GPs come equipped with the possibility to obtain a predictive variance (i.e. error bars, confidence intervals) for each prediction. Unfortunately, the GP formulation usually assumes that there is no noise in the inputs, only in the observations. However, this is often not the case in Earth observation problems where an accurate assessment of the measuring instrument error is typically available, and where there is huge interest in characterizing the error propagation through the processing pipeline.  In this paper, we demonstrate how one can account for input noise estimates using a GP model formulation which propagates the error terms using the derivative of the predictive mean function. We analyze the resulting predictive variance term and show how they more accurately represent the model error in a temperature prediction problem from infrared sounding data.
\end{abstract}


\maketitle

\section{Introduction}
	\label{sec:intro}
	The land and sea surface temperature of the Earth is one of the most important components to understanding the governing physical processes on the Earth's system~\cite{blondeau2014}. Derived processes such as heat-fluxes and energy balances on a large temporal and spatial scale are useful for applications within climate change, vegetation monitoring and other environmental studies. In order to acquire a complete model for many of these applications, one needs a good characterization of temperature on a global scale~\cite{li2013}. Acquiring ground measurements is not always practical at such a high spatial and temporal resolution scale. Remote sensing has proven to be useful to collect input data for models that capture temperature and other important environmental factors. 
Instruments such as the Infrared Atmospheric Sounding Interferometer (IASI)~\cite{iasi2012}, have an objective to support numerical weather prediction (NWP) models to provide high quality predictions for temperature, humidity, and some trace gases and generating global maps from satellite acquisitions which require fast and accurate algorithms. 



A convenient complement to physical or numerical model inversion consists of using machine learning algorithms to substitute or emulate one or more parts in the processing pipeline. Gaussian Processes (GP) are examples of non-parametric regression models that have grown in popularity over the past decade~\cite{Rasmussen2005}. Their strength comes from the use of Bayesian statistics in order to produce mean predictions with confidence intervals. 
In recent years, GPs have been successful for modeling inputs and outputs on a wide variety of tasks in remote sensing and geosciences~\cite{Camps-Valls2016a}. 

However, one crucial limitation of many machine learning algorithms in general (including GPs) is their ability to handle noisy inputs. Although this is rarely studied in the machine learning community, this relationship is very important in the Earth science and remote sensing in particular. It is customary to estimate a function $f(\x)$ given some noisy observations at $\y$ for some input sample/location $\x$. Many machine learning models assume that the inputs $\x$ are noise-free and tailor their training procedure around this assumption. 
For some applications this is a valid assumption but, as the number of data points increases and originate from different sources including other models, both assumptions become invalid and this can lead to poor modeling performance and misleading conclusions in error and uncertainty propagation studies. 
Quantifying uncertainty in machine learning (ML) models is becoming more and more prevalent, despite this already being an essential part of the physical model pipeline process. On top of this, as more and more ML algorithms are being used to replace physical models, the proper analysis of the error propagation through the whole model becomes imperative. 



In this paper, we apply a GP regression formulation which allows one to account for input noise estimates by exploiting the derivative of the predictive mean function. By using this method which propagates the input error, we demonstrate that the uncertainty estimates are more credible and more accurately estimate the residual error. Section II reviews the current literature on uncertain inputs in relation to GPs and introduces the proposed predictive variance estimate. 
Section III gives the experimental results using Infrared atmospheric sounding interferometer (IASI) data to predict surface temperature. 
Finally, Section V concludes this letter and offers some possible extensions.


\section{Gaussian Processes with Noisy Inputs}
	\label{sec:gp}
	
This section introduces the problem of dealing with noisy inputs, reviews the theory of GP regression and introduces the model we use to account for the error in the inputs.

\subsection{Regression with Uncertain Inputs}
    \label{sec:theory}

\begin{figure}[h!]

\centering

\tikzstyle{true} = [rectangle, rounded corners, minimum width=3cm, minimum height=1cm,text centered, draw=black, fill=blue!30]
\tikzstyle{observed} = [rectangle, rounded corners, minimum width=4cm, minimum height=1cm,text centered, draw=black, fill=red!30]
\tikzstyle{output} = [rectangle, rounded corners, minimum width=3cm, minimum height=1cm,text centered, draw=black, fill=green!30]
\tikzstyle{arrow} = [thick,->,>=stealth]
\tikzstyle{arrowrelation} = [thick, --, >=stealth]

\begin{tikzpicture}[scale=.75,transform shape]

\centering

\node (x)  [true, minimum size=2mm] {${\bf x}$};
\node (xbar) [observed, right of=x, xshift=20mm, minimum size=2mm] {$\tilde{{\bf x}} = {\bf x} + \boldsymbol{\epsilon}_x$};
\node (y) [output,below of=x, yshift=-20mm, minimum size=2mm] {$y$};
\node (ybar) [output, below of=xbar, yshift=-20mm, minimum size=2mm] {$\hat{y}$ };

\draw [arrow] (x) --node[anchor=east] {} (y);
\draw [arrow] (x) --node[anchor=west] {$f({\bf x})$} (ybar);
\draw [arrow] (xbar) --node[anchor=west] {$f(\tilde{{\bf x}})$} (ybar);
\draw [arrow] (x) --node[anchor=north] {} (xbar);
\draw [arrow] (y) --node[anchor=south] {$\boldsymbol{\epsilon}_{y}$} (ybar);
\end{tikzpicture}

\caption{Conceptual map illustrating the relation between the noisy inputs and noisy outputs for an arbitrary model. Traditional machine learning methods use the path from ${\bf x}$ to $\hat{y}$. In this paper, we investigate the use of the path from $\tilde{{\bf x}}$ to $\hat{y}$.} 
\label{fig:concept}
\end{figure}
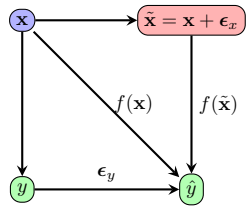

In all facets of modeling, we are essentially looking for relations between some input $\x$ and some output $y$.
Using the standard ML formulation we construct our model $f(\x)$ plus some noise. 
In GPs, this noise is actually modeled as a normal distribution with some variance $\sigma_y^2$. 
We are interested in using a GP model under the assumption that $\x$ has some noise in the inputs $\epsilon_x$ as displayed in Figure \ref{fig:concept}.

\subsection{Literature review}
In the literature, an early look at dealing with uncertain inputs is known as errors-in-variables regression \cite{Dellaportas1995}. In the more recent literature specifically related to GPs, we can divide the field into two families. The first family of methods adds complexity to the output noise model. This transforms the noise variance from a single scalar parameter, $\sigma_y^2$ to a more complex parameter that varies with respect to the output residuals. Using heteroscedastic noise models \cite{Kersting2007}, they model the noise directly in the training and testing phase. However, it is an approximation and does not explicitly exploit the structure of the input noise and yet adds more degrees of freedom to the learning algorithm.

The second family of methods try to improve the GP model by directly considering the input uncertainty. Many of the following methods described below have applications in dynamical problems involving time series.
They used local approximations of the test points from the posterior of GP function by using the Taylor approximation of the predictive mean and variance functions. Observations from \cite{Solak2003} note that derivatives of GPs are also GPs so the derivatives have a closed-form mean and variance function. However, their approach resulted in a minimum of second order derivatives of the covariance function and third order to calculate the gradients for minimizing the maximum likelihood. This adds a significant layer of model and computational complexity which is problematic with large scale problems. Another problem with this approach is that integrating over all possible trial points may not result in a Gaussian distribution. One can assume the resulting distribution of the GP model to be Gaussian and compute its mean and variance by using the approximate moments approach. This was started by \cite{Girard2003a, Quinonero-Candela2003} and further developed in \cite{Deisenroth2009, Deisenroth2011}. 
These approximations assume some noise in the $\x$ inputs but only take them into account during the predictions. Authors in \cite{Dallaire2011} apply an uncertainty incorporating covariance function used in the training phase and similarly in \cite{Girard2003}. But this case is only applicable if the error in the inputs are explicitly known and does not make adjustments to the model from the posterior information. The Noise-Input Gaussian Process (NIGP) method \cite{Mchutchon2011} constructs a GP framework that takes into account the posterior data by using a combination of the gradient of the predictive variance function and the input error covariance matrix. This method processes the input noise through the Taylor expansion and adds a corrective term which includes the derivative.

%
%
%
In remote sensing applications, typically we have well characterized measurement errors by way of sensor error estimates during the design phase. This eliminates the need to model the input error which alleviates the computational cost of finding the value of these parameters. Although all of the above methods incorporate input noise into the GP model, the ideal situation is to account for input noise in the training procedure as done in the NIGP. However training these models is computationally infeasible when dealing with a large amount of  multidimensional  data,  such  as  in  remote  sensing.  In  the following section, we introduce and analyze a simple way to better  estimate  the  error  propagation  of the test points.  The  approach  is  inspired  on  the  NIGP  formulation but adapted from the GPs used in the dynamical systems framework to suit remote sensing applications.


\subsection{Classical GP}

Let us fix the notation and the classical GP formulation first. We are given $N$ pairs of input-output points, $\{\x_i,y_i|i=1,\ldots,N\}$, where $\x=[x^1,\ldots,x^D]^\intercal\in\mathbb{R}^{D\times 1}$ and $y\in\mathbb{R}$. Let us define $\mathbf{X} =\left[\x_1|\cdots| \x_n \right]^\intercal\in\mathbb{R}^{N \times D}$ be a set of known $N$ data points, and $\mathbf{y}=\left[y_1, y_2, \ldots, y_n \right]^\intercal\in\mathbb{R}^{N\times 1}$ be the known $N$ labels in $\mathbb{R}^{N \times 1}$. We are interested in finding a latent function $f(\x)$ of input $\x$ that approximates $y$. We assume the function $f(\cdot)$ is corrupted by some noise: 
\begin{equation}\label{eq:latent}
    y = f(\x) + \epsilon_y,
\end{equation}
where $\epsilon_y$ represents the modeling error or residuals. By assuming a Gaussian prior for the noise term $\epsilon_y\sim\mathcal{N}(0, \sigma^2_y)$, and a zero mean GP prior for the latent function, $f(\x) \sim \mathcal{GP} \left( \mathbf{0}, \mathbf{K} \right)$, where $\mathbf{K}$ is the covariance matrix parameterized using a kernel function, $\mathbf{K}_{ij} = k(\x_i,\x_j)$, we can analytically compute the posterior distribution over the unknown output $y_*$,
with the following predictive mean and variance for a new incoming test input point $\x_*$: 
\begin{eqnarray}
	\mathbb{\mu}_{GP} &=& {\bf k}^\intercal_* ({\bf K}+\lambda {\bf I}_N)^{-1}{\bf y} = {\bf k}^\intercal_* \alpha \label{eq:gp_pred} \\
    \mathbb{\nu}^2_{GP} &=& k_{**}- {\bf k}^\intercal_* ({\bf K}+\sigma_y^2 \mathbf{I}_N)^{-1} {\bf k}_{*},
    \label{eq:gp_var}
\end{eqnarray}
where ${\bf k}_* = \left[k(\x_*, \x_1), k(\x_*, \x_2), \ldots, k(\x_*, \x_N) \right]^{\intercal} \in \mathbb{R}^{N\times 1}$ calculates the similarities between the test point $\x_*$ and all of the training samples, $k_{**}=k(\x_*,\x_*)$ is the self-similarity matrix for the test samples, and $\mathbf{I}_N$ is an identity matrix of size $N \times N$.


\subsection{GP Regression with Noisy Inputs}
To account for noisy inputs, we can restart the GP formulation under the assumption that $\tilde{\x}$ is the real vector that contains the real observed input $\x$ corrupted by some $\epsilon_x$ as in figure \ref{fig:concept}.
By introducing $\tilde{\x}$ in Eq. \eqref{eq:latent}, we can obtain the following model which includes the input noise in the latent function $f(\tilde{\x})$ which is corrupted by two sources of noise $\epsilon_y$ and $\epsilon_x$,
\begin{equation*}
    y = f(\x + \epsilon_x) + \epsilon_y,
\end{equation*}

\begin{figure}
\begin{center}
\setlength{\tabcolsep}{1pt}
\begin{tabular}{ccc}
(a) & (b) & (c) \\[-0.1cm]
\includegraphics[height=30mm,width=30mm,trim={2cm 1cm 2cm 1cm},clip]{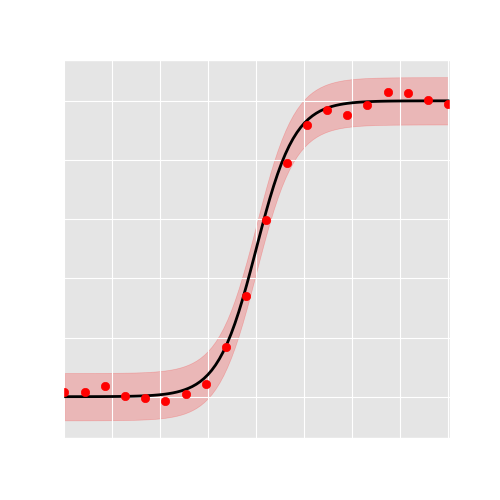} &
\includegraphics[height=30mm,width=30mm,trim={2cm 1cm 2cm 1cm},clip]{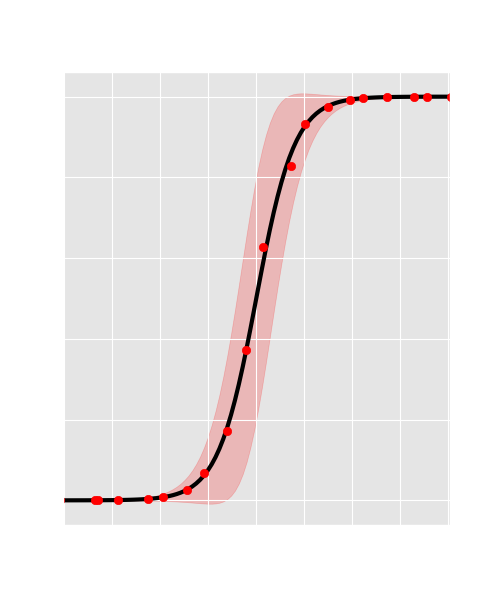} &
\includegraphics[height=30mm,width=30mm,trim={2cm 1cm 2cm 1cm},clip]{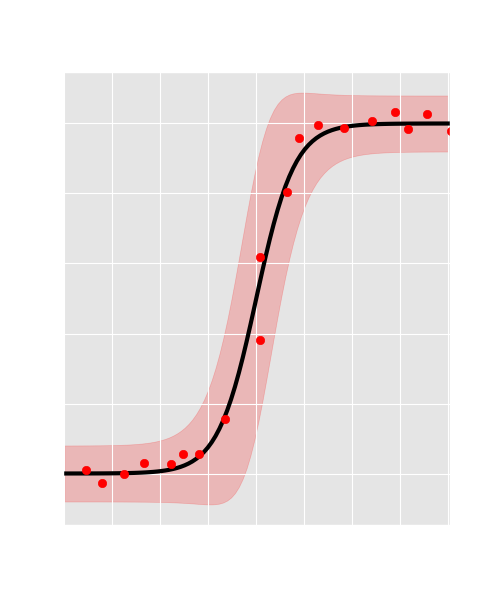}
\\
\end{tabular}
\vspace{-0.15cm}
\caption{Effects of considering (a) only noise for the outputs $y$, (b) only noise for the inputs $x$, and (c) noise for both the inputs $x$ and the outputs $y$.} \label{fig:derivative}
\end{center}
\end{figure}

 The (first-order) Taylor expansion centered at $\x$ provides us with a similar formulation:
%
\begin{equation*}
y \approx f(\x) + \epsilon_x^\intercal\partial_{\bar{f}} + \epsilon_y. 
\label{eq:error}
\end{equation*}
This result gives us insight into the {\em error propagation} of the input noise by using the derivatives of the mean function. One may imagine that in regions where there is a high gradient (where the output value is rapidly changing), the input measurements corrupted by noise will be more important as opposed to regions where the output is almost constant with regard the input. Figure \ref{fig:derivative} illustrates this effect. 

The same Gaussian prior for the noise term $\epsilon_y$ and the zero mean GP prior for the latent function $f(\x)$ is used, like in the standard GP formulation. We can assume an additive white Gaussian noise for the inputs, $\epsilon_x \sim \mathcal{N}(0, \boldsymbol{\Sigma}_x)$. Now we have a typical problem since we cannot analytically compute the posterior distribution of the unknown output because now our $\x$ comes from a distribution itself, $\x\sim \mathcal{N}(0,\boldsymbol{\Sigma}_x)$ \cite{Mchutchon2014} resulting in a non-Gaussian distribution. However we can simply take the expectation and variance of our new function~\cite{Mchutchon2011,Mchutchon2014} to be approximated as a Guassian. The expectation gives us the same sample GP prior mean, but the resulting equation for the variance of the unknown outputs $y_*$ for a new incoming test input point $\x_*$ changes as:
\begin{equation}
    \mathbb{\nu}^2_{eGP} = {T}_{**} + k_{**}- {\bf k}^\intercal_* ({\bf K}+\sigma_y^2 \mathbf{I}_N + {\bf T})^{-1} {\bf k}_{*}
    \label{eq:gp_var_unc}
\end{equation}

where the effect of the noise in the inputs is represented by ${\bf T}_{ij} = T(\x_i,\x_j) = \partial_{i}^\intercal\boldsymbol{\boldsymbol{\Sigma}}_x\partial_{j}$, and $T_{**} = T(\x_*,\x_*)$. We denoted the vector of partial derivatives of $f$ w.r.t. the sample $x_i$ as:

\begin{equation*}
\partial_{i}:=\Big[\frac{\partial f(\x_i)}{\partial x_i^1}\dots\frac{\partial f(\x_i)}{\partial x_i^D}\Big]^\intercal. 
\end{equation*}

The derivative of the predictive function $f$ (eq.  \eqref{eq:gp_pred}) in GPs only depends on the derivative of the kernel function since it is linear with respect to the $\boldsymbol{\alpha}$ parameters:
\begin{align*}\label{eq:der}
	\dfrac{\partial f(\x_i)}{\partial x_i^j} = 
    \dfrac{\partial {\bf k}_i \boldsymbol{\alpha}}{\partial x_i^j} 
    = (\partial {\bf k}_{ij})^\intercal \boldsymbol{\alpha},
\end{align*} %

where $\partial {\bf k}_{ij} = [\frac{\partial k(\x_i,\x_1)}{\partial x_i^j},\ldots,\frac{\partial k(\x_i,\x_N)}{\partial x_i^j}]^\intercal$. Please see \footnote{\url{https://isp.uv.es/egp.html}} for a working implementation of the error GP (eGP) model.

\section{Experiments}
	\label{sec:exp}
	

\subsection{An illustrative example}\label{sec:toy}

\begin{figure}[t!]
\begin{center}
\begin{tabular}{cc}
\includegraphics[height=3cm,width=4cm,trim={2cm 1cm 2cm 1cm},clip]{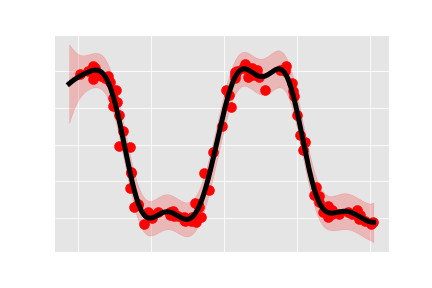} &
\includegraphics[height=3cm,width=4cm,trim={2cm 1cm 2cm 1cm},clip]{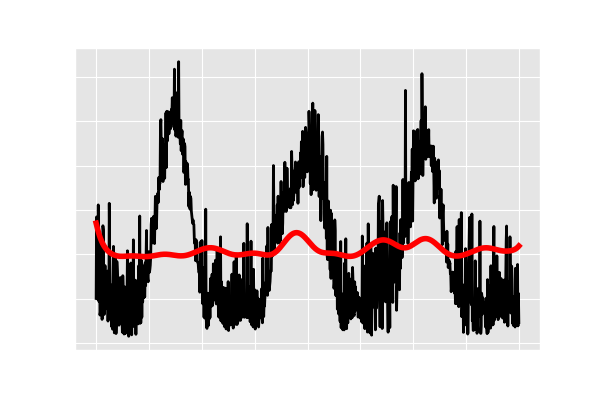}
\\
\includegraphics[height=3cm,width=4cm,trim={2cm 1cm 2cm 1cm},clip]{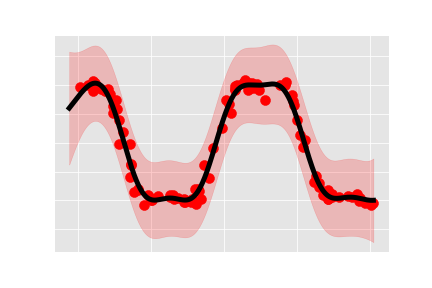} &
\includegraphics[height=3cm,width=4cm,trim={2cm 1cm 2cm 1cm},clip]{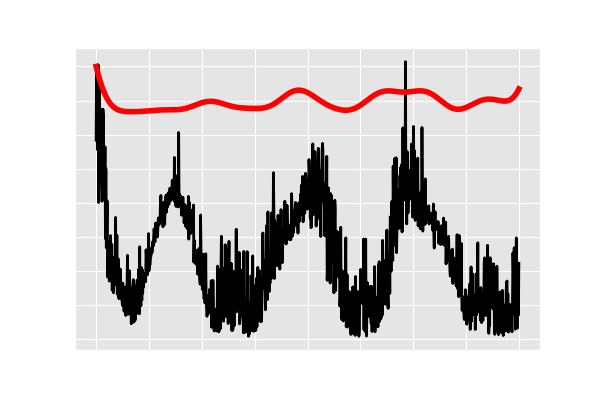}
\\
\includegraphics[height=3cm,width=4cm,trim={2cm 1cm 2cm 1cm},clip]{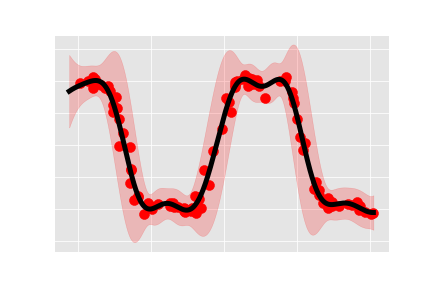}  &
\includegraphics[height=3cm,width=4cm,trim={2cm 1cm 2cm 1cm},clip]{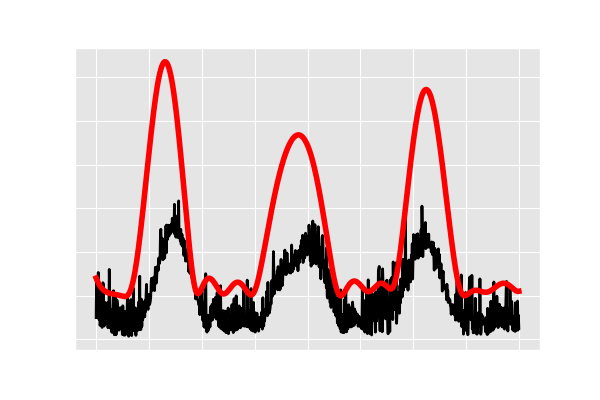}
\\
\end{tabular}
\vspace{-0.25cm}
\caption{Illustrative uni-dimensional example showing the predictive mean and predictive variance using different GP methods: standard GP (top), GP with heteroscedastic noise levels (middle) and GP with variance moment matching error propagation (bottom). The left column shows the predictive mean and the predictive variance for each method. The right column shows the empirical variance (red line) in relation to the empirical predictive variance (black line). An input noise coefficient of $\sigma_x=0.3$ and an output noise coefficient $\sigma^2_y=0.05$ was used for this demonstration. 
} 
\label{fig:1dexp}
\end{center}
\end{figure}

Figure \ref{fig:1dexp} showcases a simple ``nearly-square'' sine wave and how three different GP formulations approximate the variance of the function. 
A standard GP (top row) clearly does not have any correspondence to the errors in the inputs as the line is nearly constant for all regions. A GP using a heteroscedastic noise model (middle row) does capture and adjust for the input noise but the output still remains constant with no confidence in regions where the outputs do not vary. On the bottom row, the GP with our adjusted variance correctly adjusts the predictive variance based on regions where the gradient is higher and the outputs drastically change. This same response to the input noise is what we are hoping to accomplish with a real and more complex dataset in the following sections \ref{subsec:data}.

\subsection{Temperature estimation from infrared sounding data}\label{subsec:data}

We illustrate how the proposed predicted variance accounting for the input errors (eGP) compares to the standard predictive variance for typical GP models when trained to estimate surface temperature from noisy input radiance values. 

\subsubsection{Data}

We use data acquired by the IASI instrument onboard the MetOp-A satellite, which consists of 8461 spectral channels between 3.62 and 15.5 $\mu$m with a spectral sampling of 0.25 cm$^{-1}$ and a spatial resolution of 25km. We chose the 01-10-2013 for our sample space, which contained $13$ complete orbits within a $24$-hour period. Since temperature is an exemplary atmospheric parameter for weather forecasting, we used the atmospheric surface temperature predictions of the European Centre for Medium-Range Weather Forecasts (ECMWF) model and the radiance IASI measurements. The IASI instrument is well characterized and the noise can be described as an additive Gaussian noise with a covariance error for the radiance values provided by European Organization for the Exploitation of Meteorological Satellites (EUMETSAT).

\subsubsection{Methodology}

We have 13 orbits in total. $N_{\text{train}}=5000$ points were selected as training samples randomly from all orbits. The extremely noisy channels were discarded to reduce the dimensionality of the data from $8461$ to 4699. This was followed by Principal Component Analysis (PCA) to further reduce the number of dimensions from over 4699 to $50$ accounting for $99\%$ of the variance within the data. Using this reduced sample space, we train a standard GP model using a negative maximum-log-likelihood scheme with $10$ random optimizer restarts for a standard Radial Basis Function (RBF) kernel. The remaining points; total of $N_{\text{test}}=1,182,600$ points; were used for testing. We use the same standard GP to calculate the predictive mean of for the test points. We calculate the standard standard deviation (eq. \eqref{eq:gp_var}) and our augmented standard deviation with input variances (eq. \eqref{eq:gp_var_unc}) to compare to the mean absolute error.

\subsubsection{Temperature estimation}

\begin{figure}[t!]
\begin{center}
\vspace{-3mm}

\begin{tabular}{c}
\\
\hspace{-1.3cm}
(a) Ground Truth
\vspace{-1.0cm}
\\
\hspace{-1.3cm}
\includegraphics[height=7cm,width=10.5cm]{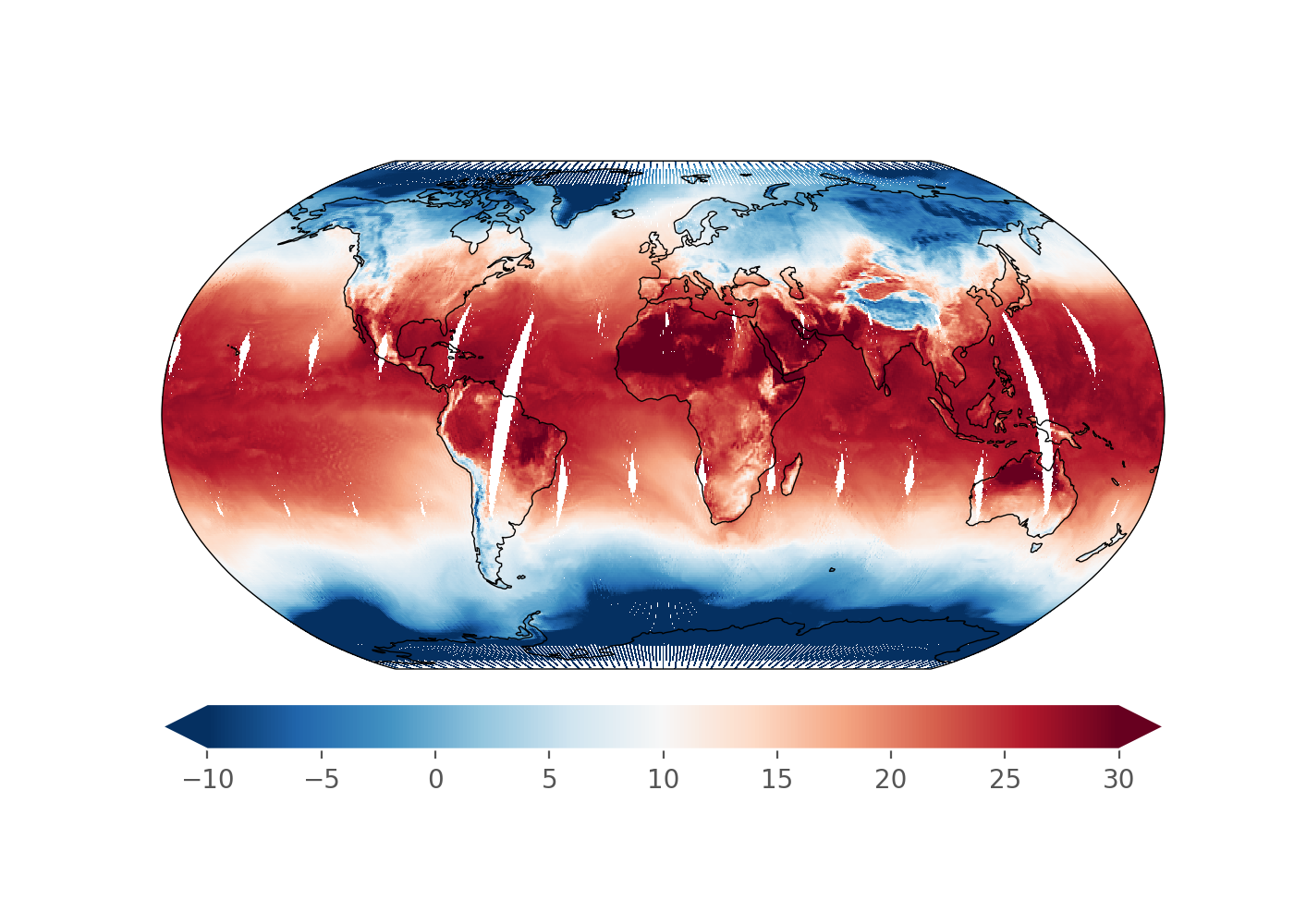}
\vspace{-8mm}
\\
\hspace{-1.3cm}
(b) Predictions
\vspace{-1.0cm}
\\
\hspace{-1.3cm}
\includegraphics[height=7cm,width=10.5cm]{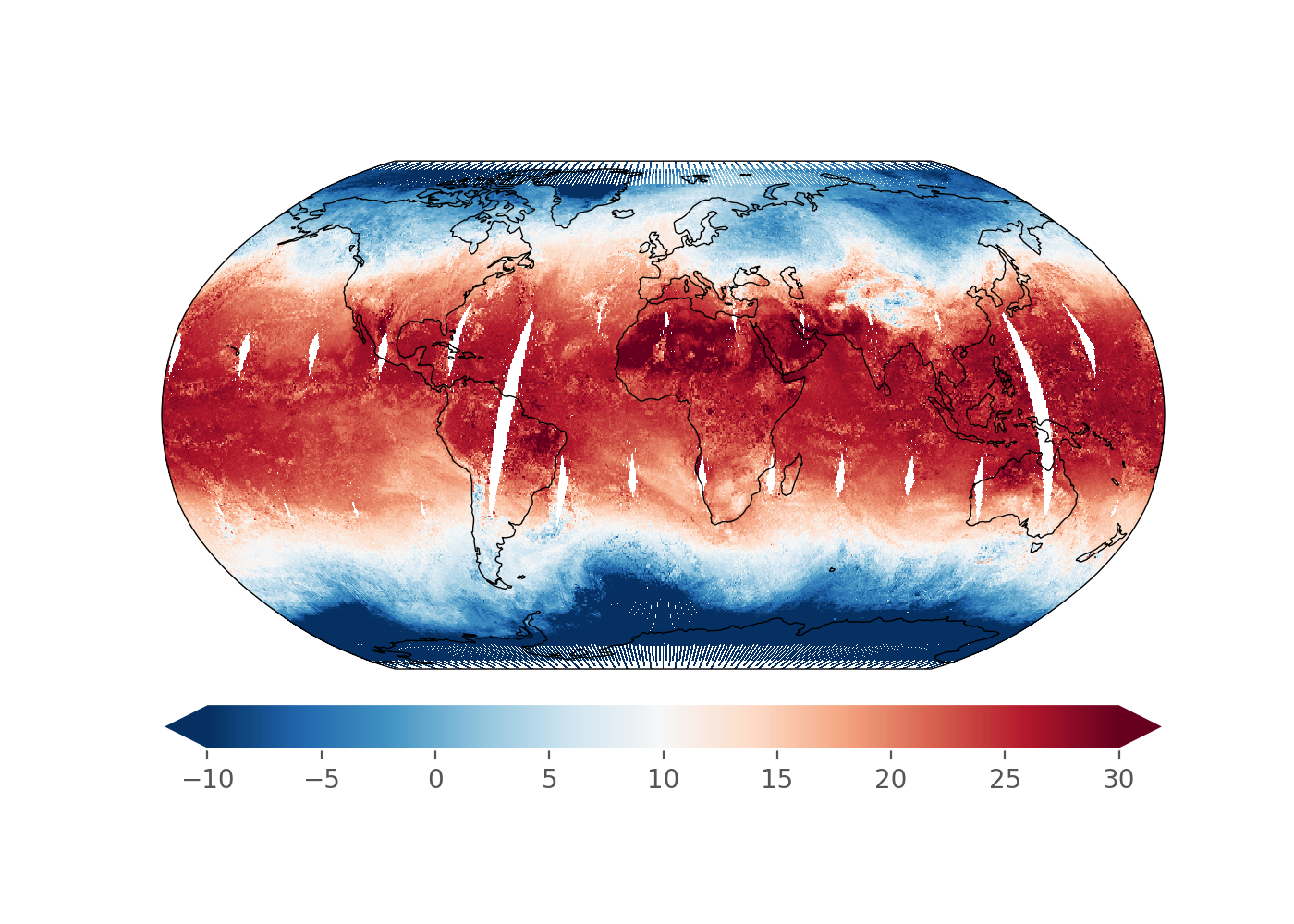}
\vspace{-8mm}

\end{tabular}
\vspace{-0.0cm}
\caption{Mean temperature prediction for 01-10-2013. (a) shows the ground truth from the ECMWF model and (b) shows our predictions using the predictive mean of the GP.} 
\label{fig:results}
\end{center}
\end{figure}

Statistically, the estimation using the predictive mean of the trained GP model achieved an average mean absolute error $e_{GP}=|y- \mu_{GP}|$ of around 2$^\circ$C and a model R$^2$ value of 0.97.
The first set of temperature maps in Fig. \ref{fig:results} show the mean surface temperature ground truth provided by ECMWF model versus the GP model predictions. Visibly, the results are similar but there are some discrepancies in regions where there is a large change in temperature. For example regions along the boundaries between the red and blue in the southern and northern regions of the equation exhibit errors in predictions. Furthermore, the boundary along the west coast of south America, regions near the equator and tropic of Capricorn, and the region in central Asia have different temperature predictions than the ground truth. Sub-figure (\ref{fig:error}a) supports these visible observations as the reddest regions are the same, and also has red-regions along the northern and southern hemispheres. 

\subsubsection{Error propagation}
\begin{figure}[t!]
\begin{center}
\vspace{3mm}
\begin{tabular}{c}
\hspace{-1.3cm}
(a) Absolute Error, $e_{GP}$
\vspace{-1.0cm}
\\
\hspace{-1.3cm}
\includegraphics[height=7cm,width=10.5cm]{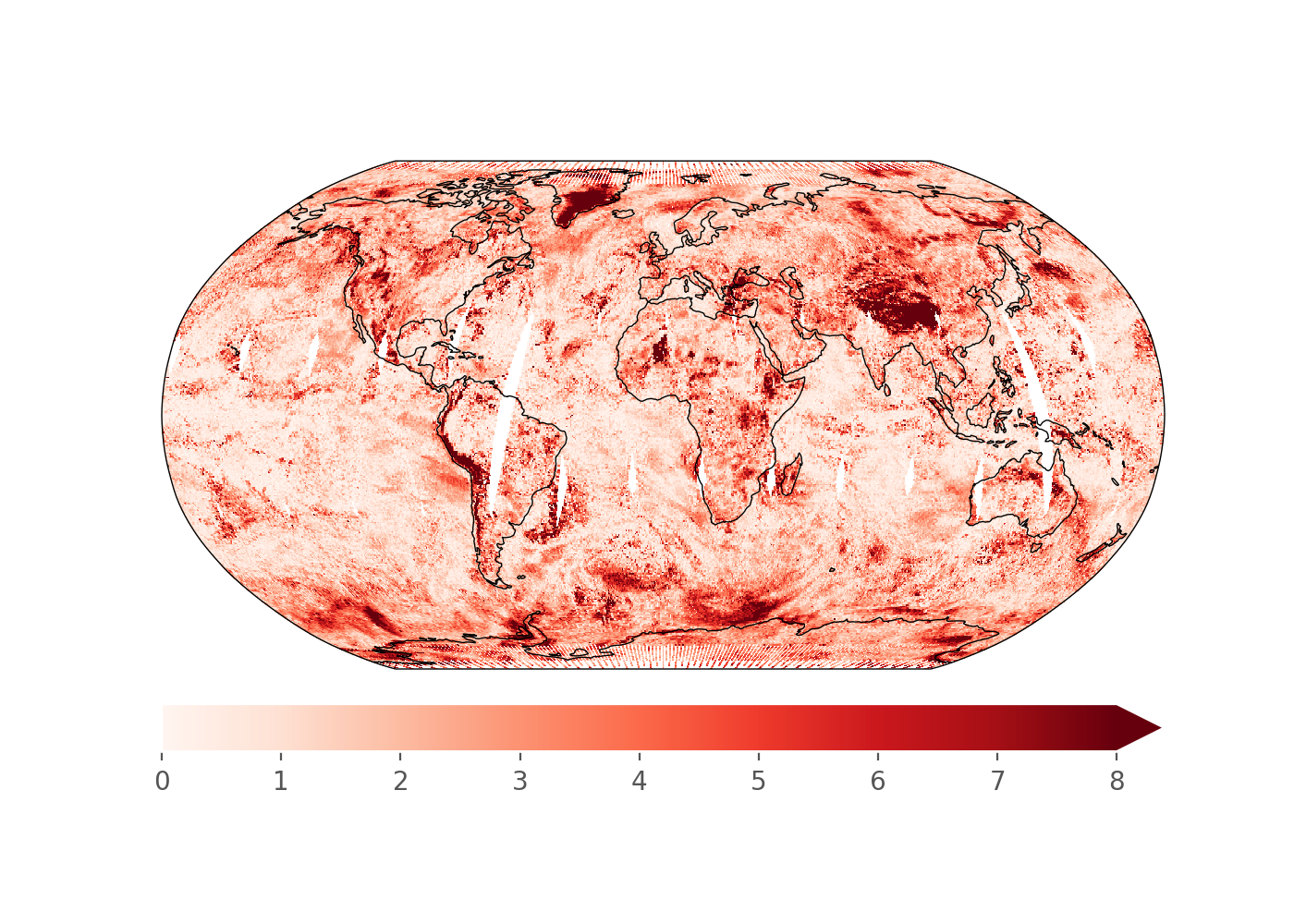}
\vspace{-8mm}
\\
\hspace{-1.3cm}
(b) Standard Deviation, $\nu^2_{GP}$
\vspace{-1.0cm}
\\
\hspace{-1.3cm}
\includegraphics[height=7cm,width=10.5cm]{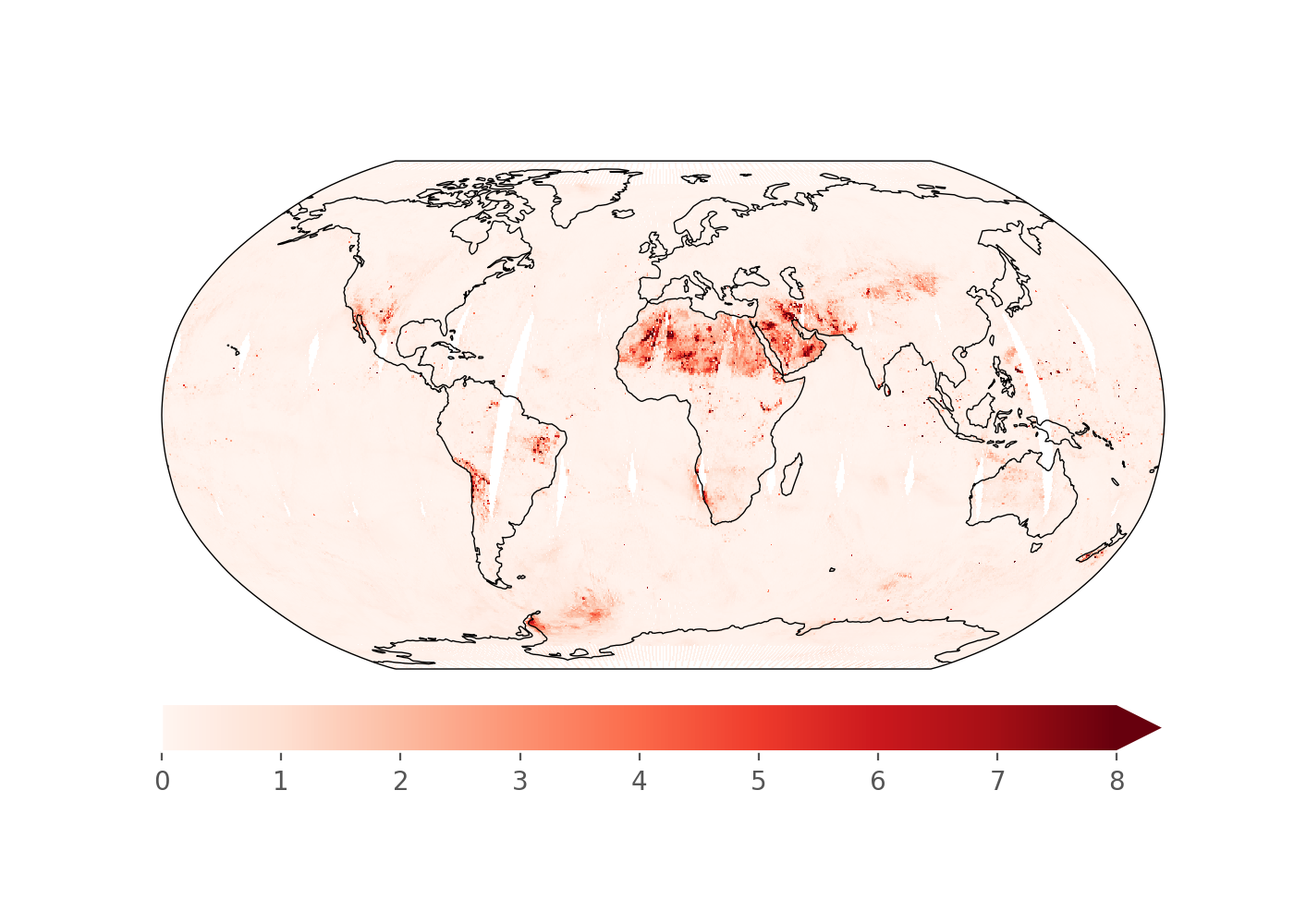}
\vspace{-8mm}
\\
\hspace{-1.3cm}
(c) Standard Deviation, $\nu^2_{eGP}$
\vspace{-1.0cm}
\\
\hspace{-1.3cm}
\includegraphics[height=7cm,width=10.5cm]{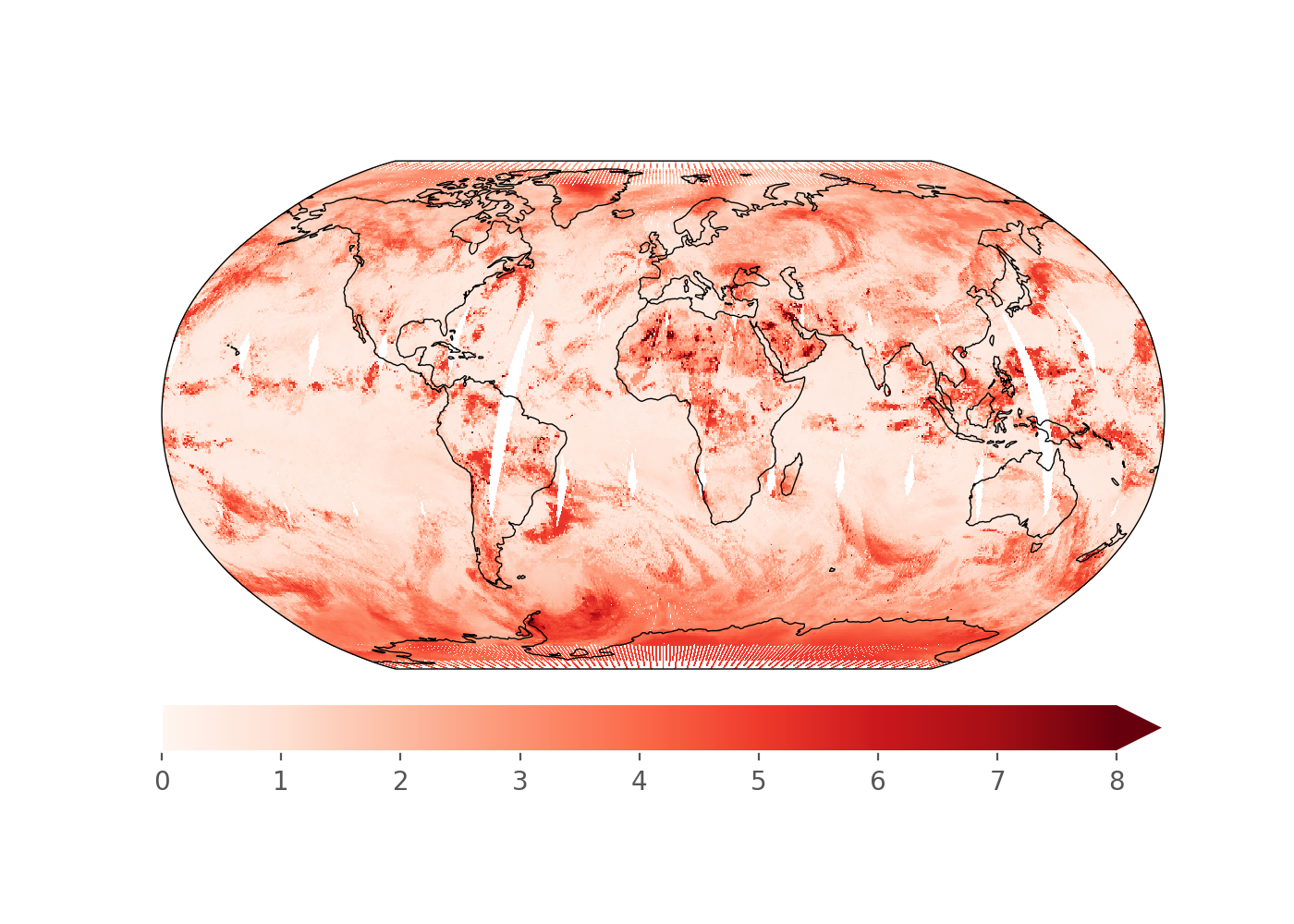}
\vspace{-8mm}

\end{tabular}
\vspace{-0.0cm}
\caption{Errors and standard deviations of our GP model for the mean temperature with orbit 01-10-2013. (a) absolute error between our predictions and the ECMWF model, (b) standard deviation from a regular GP (eq. \eqref{eq:gp_var}), and (c) proposed standard deviation which takes into account the noise in the inputs, eGP (eq. \eqref{eq:gp_var_unc}).} 
\label{fig:error}
\end{center}
\end{figure}

The maps for the two GP variance predictors are noticeably different (sub-figures \ref{fig:error}b, \ref{fig:error}c). The standard GP variance focuses on the region in northern Africa and the middle east. This is an absent region in the error map. This is also a known hot region which is expected to have a high temperature gradient locally but relatively similar temperature gradient spatially. The standard deviation for the eGP also has a large confidence interval for this area but in addition chooses regions where there is a high temperature gradient spatially; like the aforementioned regions in the norther and southern hemispheres. Many of the slightly less red regions correspond to the regions on the error map. Overall, it appears that the spread between the regions with lower and higher standard deviations are more pronounced in the eGP than the standard GP which has a high concentration on regions where the temperature is spatially similar.

%
%
%
Table \ref{tb:corr} shows the numerical results. It presents the error estimates between the standard deviation for the GP and eGP versus the mean absolute error between the predictions and the labels. In all cases, the eGP and the absolute error has lower error statistics. 
\begin{table}[]
\caption{
Metrics between the \textbf{mean absolute error} ($e_{GP}$) of the GP model and the \textbf{predictive standard deviation} for both the GP $\nu^2_{GP}$ (\ref{eq:gp_var}) and eGP $\nu^2_{eGP}$ (\ref{eq:gp_var_unc}).} 
\centering
    \begin{tabular}[t]{|l|c|c|c|c|}
    \hline\hline
    \multicolumn{1}{|c|}{\multirow{2}{*}{Statistic}} & \multicolumn{2}{c|}{$\nu^2_{GP}$} & \multicolumn{2}{c|}{$\nu^2_{eGP}$}  \rule{0pt}{2.5ex}\rule[-1.2ex]{0pt}{0pt}\\ 
    \cline{2-5} 
    \multicolumn{1}{|c|}{}                           & Mean            & Variance          & Mean                & Variance            \\ \hline
    Mean Absolute Error                              & 1.92            & 3.05              & \textbf{1.20}       & \textbf{1.63}       \\ \hline
    Mean Squared Error                               & 7.90            & 81.91             & \textbf{4.10}       & \textbf{80.85}      \\ \hline
    Root Mean Squared Error                          & 2.81            & 9.05              & \textbf{2.05}       & \textbf{9.00}       \\ \hline\hline
    \end{tabular}
    \label{tb:corr}
\end{table}
	
\section{Conclusion}
	\label{sec:conc}
	The consideration of noisy inputs is extremely important in Earth science for error characterization and uncertainty quantification and propagation. However, their formal treatment in machine learning has not been widely approached. If we hope to combine the use of statistical models with physical models, then we will need accurate error and uncertainty estimates for our predictions. 

In this paper, we gave a simple formulation and rationale for how the derivative of GP models in particular can be used to help the predictive variance obtain more accurate (and credible) error estimates in Earth science applications. 
Using a GP model to predict temperature from radiances, we showed quantitatively and visually that the predictive variance with the propagated error provided a stronger correspondence to the absolute error and that it can be useful to understand a GP models performance and noise/errors impact. 

For further work, one could incorporate the input noise information both during the training procedure of the GP algorithm and in the computation of the predictive variance in the testing procedure as well as incorporate this same framework of utilizing the derivatives of kernel functions to propagate the error in other kernel methods.

\section{Acknowledgements}

This research was funded by the European Research Council (ERC) under the ERC-Consolidator Grant 2014 Statistical Learning for Earth Observation Data Analysis. project (grant
agreement 647423).
\bibliographystyle{bibliographies/IEEEtran}
\bibliography{bibliographies/IEEEabrv, bibliographies/gp_error}

\end{document}